\documentclass[final]{cvpr}

\usepackage{times}
\usepackage{epsfig}
\usepackage{graphicx}
\usepackage{amsmath}
\usepackage{amssymb}

\usepackage[pagebackref=true,breaklinks=true,colorlinks,bookmarks=false]{hyperref}
\usepackage{bbm}
\usepackage[utf8]{inputenc} 
\usepackage{pifont}
\usepackage{amsfonts,amssymb,nicefrac,array,mathtools}     
\usepackage{microtype}
\usepackage{graphicx,epsfig,subcaption}
\usepackage{multirow,multicol,booktabs}
\usepackage[flushleft]{threeparttable}
\usepackage{color,xcolor}
\usepackage{comment}
\usepackage{algorithmic}
\usepackage[ruled,vlined]{algorithm2e}
\usepackage{mwe}
\usepackage{adjustbox}
\usepackage{enumitem}

\usepackage{lipsum}

\newcommand{\tablestyle}[2]{\setlength{\tabcolsep}{#1}\renewcommand{\arraystretch}{#2}\centering\footnotesize}

\newcommand{\app}{\raise.17ex\hbox{$\scriptstyle\sim$}}

\newlength\savewidth\newcommand\shline{\noalign{\global\savewidth\arrayrulewidth
  \global\arrayrulewidth 1pt}\hline\noalign{\global\arrayrulewidth\savewidth}}
  
\makeatletter\renewcommand\paragraph{\@startsection{paragraph}{4}{\z@}
  {.5em \@plus1ex \@minus.2ex}{-.5em}{\normalfont\normalsize\bfseries}}\makeatother

\def\fig#1{Fig.~\ref{fig:#1}}

\usepackage{adjustbox}
\usepackage{graphicx}
\usepackage{amsmath}
\usepackage{colortbl}
\usepackage[switch]{lineno}

\usepackage{enumitem}
\usepackage{wrapfig}
\usepackage{lipsum}
\usepackage{xcolor}
\usepackage{tabularx}
\definecolor{ForestGreen}{rgb}{0.13, 0.55, 0.13}
\definecolor{Green}{rgb}{0.0, 0.5, 0.0}
\definecolor{green(munsell)}{rgb}{0.0, 0.66, 0.47}
\definecolor{green(ryb)}{rgb}{0.4, 0.69, 0.2}
\definecolor{green(pigment)}{rgb}{0.0, 0.65, 0.31}

\def\cvprPaperID{3445} 
\def\confYear{CVPR 2021}

\begin{document}

\title{
Unsupervised Visual Attention and Invariance for Reinforcement Learning
}

\pagenumbering{gobble}

\author{
Xudong Wang\thanks{Equal contribution.}
\hspace{15mm}
Long Lian$^*$
\hspace{15mm}
Stella X. Yu\\
UC Berkeley / ICSI\\
{\tt\small \{xdwang,longlian,stellayu\}@berkeley.edu}
}

\maketitle

\begin{abstract}
Vision-based reinforcement learning (RL) is successful, but how to generalize it to unknown test environments remains challenging.  Existing methods focus on training an RL policy that is universal to changing visual domains, whereas we focus on extracting visual foreground  that is universal, feeding clean invariant vision to the RL policy learner.  Our method is completely unsupervised, without manual annotations or access to environment internals.
   
Given videos of actions in a training environment, we learn how to extract foregrounds with unsupervised keypoint detection, followed by unsupervised visual attention to automatically generate a foreground mask per video frame.  We can then introduce artificial distractors and train a model  to reconstruct the clean foreground mask from noisy observations.  Only this  learned model is needed during test to provide distraction-free visual input to the RL policy learner.

Our Visual Attention and Invariance (VAI) method significantly outperforms the state-of-the-art on visual domain generalization, gaining 15$\sim$49\% (61$\sim$229\%) more cumulative rewards per episode on DeepMind Control (our DrawerWorld Manipulation) benchmarks.  Our results demonstrate that it is not only possible to learn domain-invariant vision without any supervision, but freeing RL from visual distractions also makes the policy more focused and thus far better.

\end{abstract} 
\def\figTeaser#1{
\begin{figure}[#1]
\centering
\includegraphics[width=1.0\linewidth]{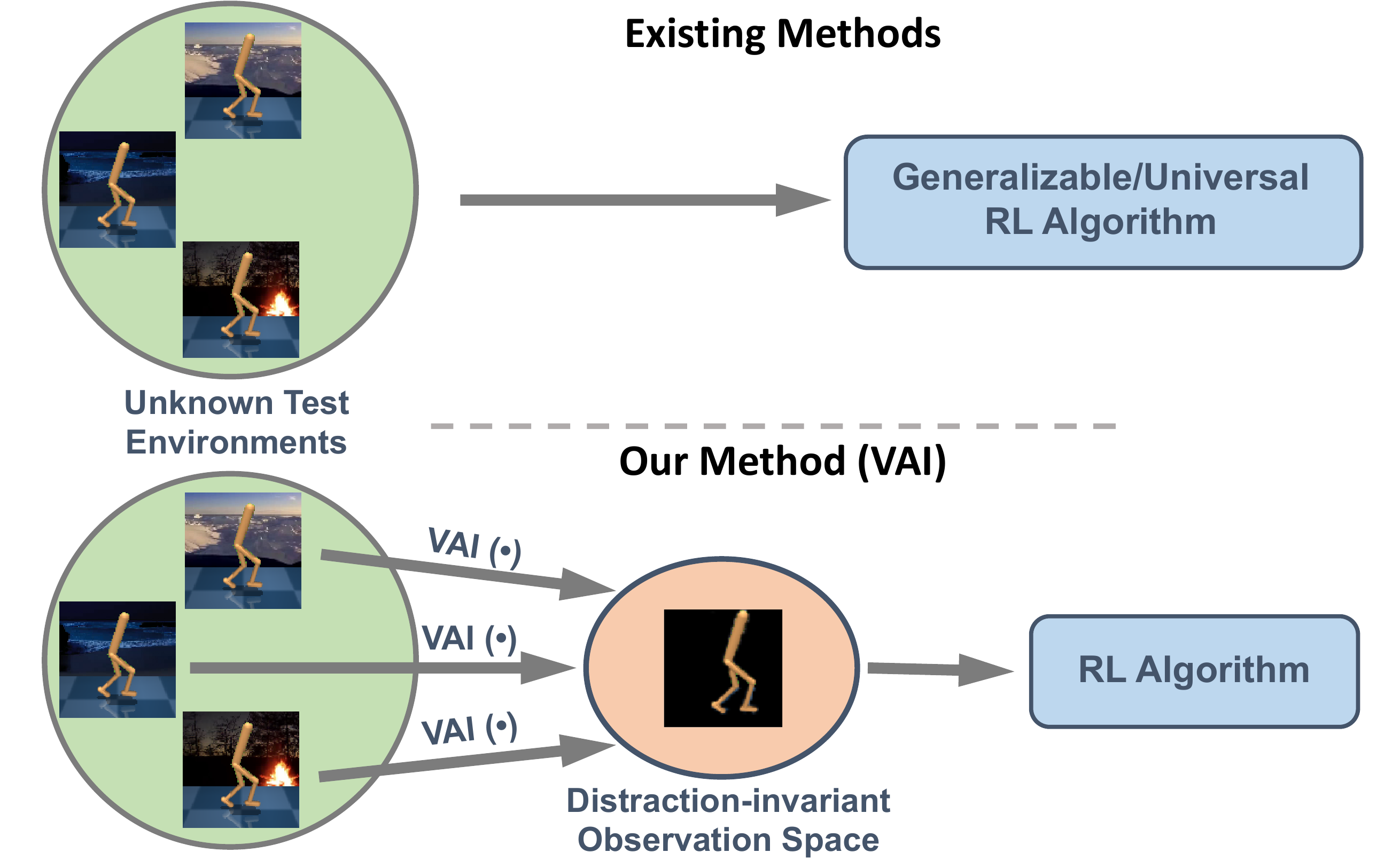}
\\[5pt]
\begin{tabular}{@{}cc@{}}
\includegraphics[width=0.49\linewidth]{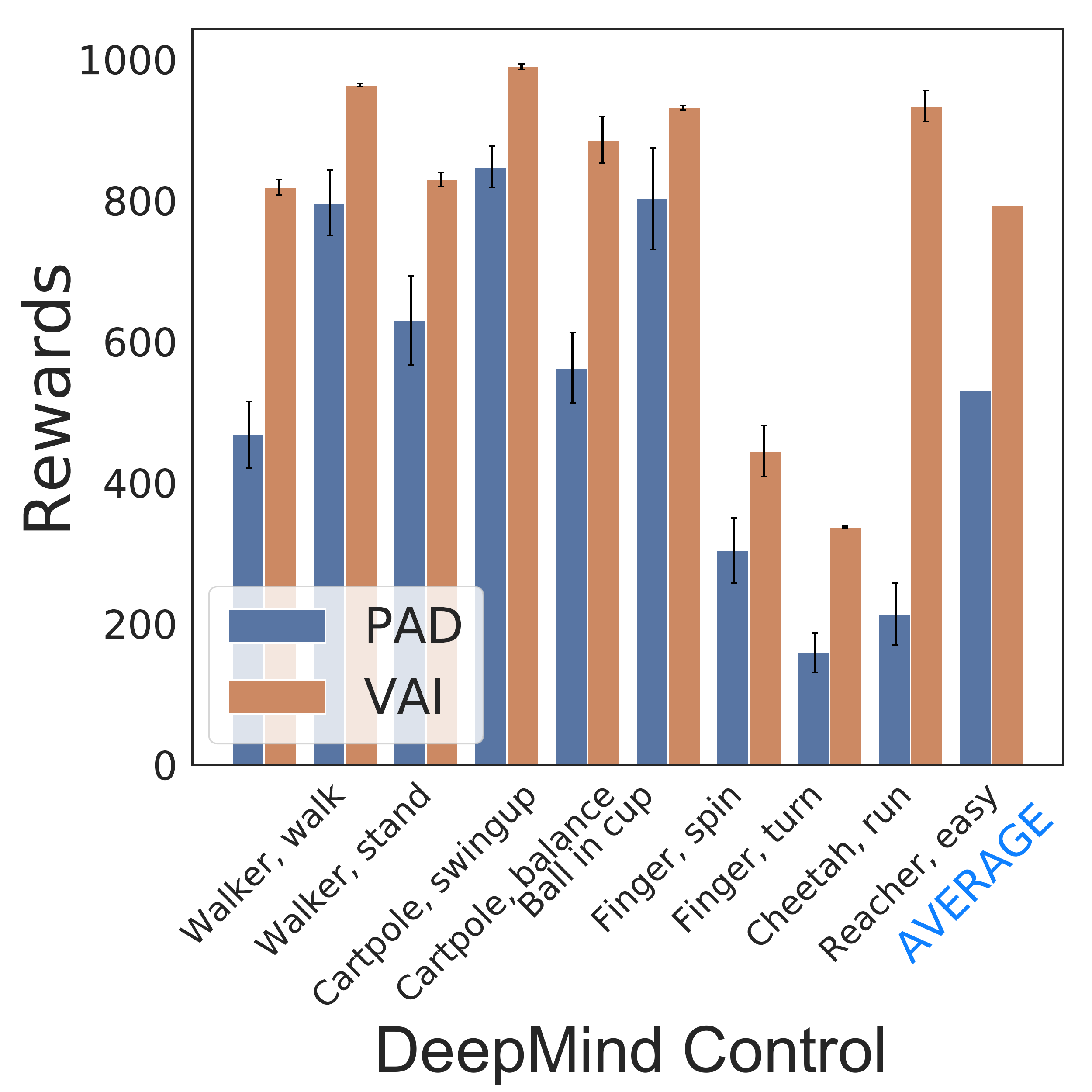}&
\includegraphics[width=0.49\linewidth]{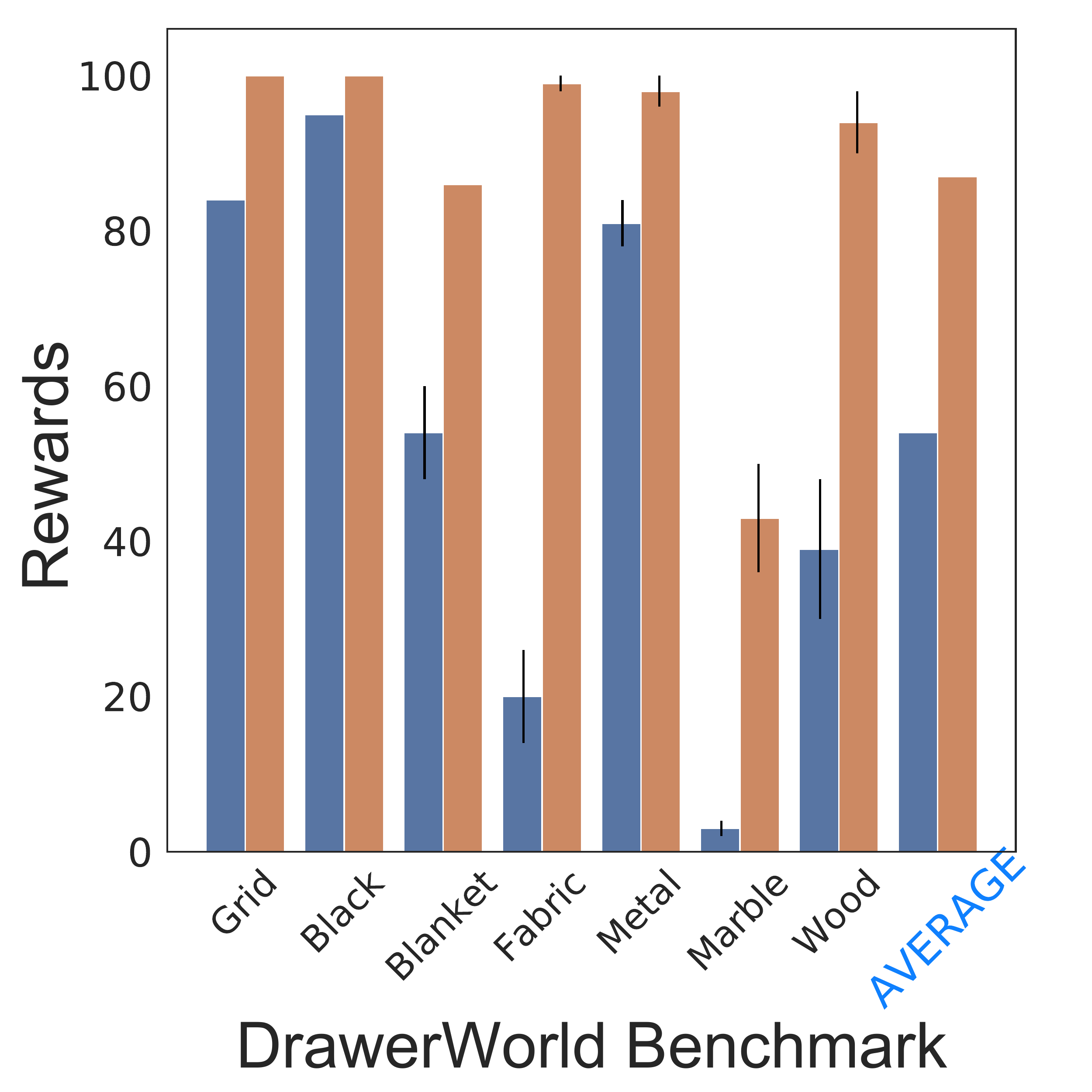}\\[-4pt]
\end{tabular}
\caption{
{\bf Top)} Two ways to make vision-based reinforcement learning generalizable to unknown environments at the test time:  Existing methods focus on learning an RL policy that is universal to varying domains, whereas our proposed Visual Attention and Invariance (VAI) extracts visual foreground  that is universal, feeding clean and invariant vision to RL. 
{\bf Bottom)} VAI significantly outperforms PAD (SOTA), increasing cumulative rewards by 49\% and 61\% respectively in random color tests on DeepMind control  and random texture tests on our DrawWorld manipulation benchmarks.
}
\label{fig:teaser}
\end{figure}
}

\def\figFrameworkSimple#1{
\begin{figure*}[#1]
\centering
\includegraphics[width=1.0\textwidth]{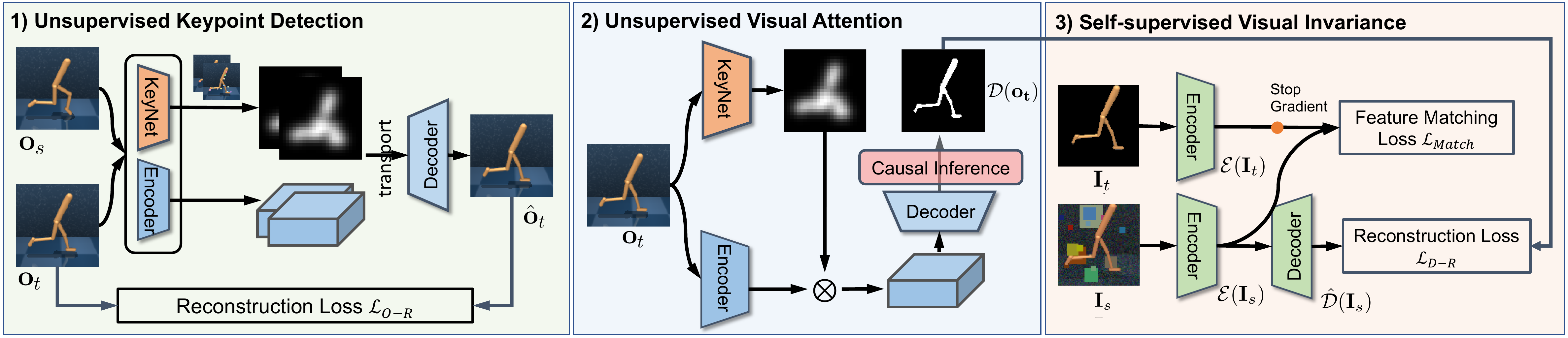}\vspace{-2pt}
\caption{Our VAI method has three components.
{\bf 1) Unsupervised keypoint detection}:  Given two adjacent video frames, we learn to predict keypoints and visual features from each image so that foreground features and target keypoints can be used to  reconstruct the target frame, without any manual annotations.
{\bf 2) Unsupervised visual attention}: We apply causal inference to remove the model bias from the foreground mask derived from detected keypoints.
{\bf 3) Self-supervised visual invariance}: We are then able to add artificial distractors and train a model to reconstruct the clean foreground observations.
Keypoint and attention modules are only used during training to extract foregrounds from videos without supervision, whereas only the last encoder/decoder (colored in green) trained for visual invariance is used to remove distractors automatically at the test time.
}
\label{fig:frameworkSimple}
\end{figure*}
}

\section{Introduction}
\figTeaser{!t}
\figFrameworkSimple{!t}

Vision-based deep reinforcement learning (RL) has achieved considerable success on robot control and manipulation.  Visual inputs provide rich information that are easy and cheap to obtain with cameras \cite{mason2001mechanics,mnih2013playing,levine2016end,finn2017deep,ebert2018visual}.  However, vision-based RL remains challenging:  It not only needs to process high-dimensional visual inputs, but it is also required to deal with significant variations in new test scenarios (Fig. \ref{fig:teaser}), e.g. color/texture changes or moving distractors \cite{nair2018visual,andrychowicz2020learning}.

One solution is to learn an ensemble of policies, each handling one type of variations \cite{rajeswaran2016epopt}.  However, anticipating all possible variations quickly becomes infeasible; domain randomization methods \cite{sadeghi2016cad2rl,tobin2017domain,peng2018sim,pinto2017robust,yang2019single} apply augmentations in a simulated environment and train a \textit{domain-agnostic universal policy} conditioned on estimated discrepancies between testing and training scenarios.

Two caveats limit the appeal of a universal RL policy. 
{\bf 1) Model complexity.}  The policy learner must have enough complexity to fit a large variety of environments.   While there are universal visual recognition and object detection models that can adapt to multiple domains \cite{rebuffi2017learning,wang2019towards}, it would be hard to accomplish the same with a RL policy network often containing only a few convolutional layers.
{\bf 2) Training instability.} 
RL training could be brittle, as gradients for (often non-differentiable) dynamic environments can only be approximated with a high variance through sampling.
Adding strong augmentations adds variance and further instability, causing inability to converge. \cite{hansen2020self} handles instability with weaker augmentations, in turn reducing generalization.

The state-of-the-art (SOTA) approach, PAD \cite{hansen2020self}, 
performs unsupervised policy adaption with a test-time auxiliary task (e.g. inverse dynamic prediction) to fine-tune the visual encoder of the policy network on the fly.  However, there is no guarantee that intermediate representations would fit the control part of the policy network.  Drastic environment changes such as background texture change from {\it grid} to {\it marble} can cause feature mismatches between adapted layers and frozen layers, resulting in high failure rates.

Instead of pursuing a policy that is universal to changing visual domains, we propose to extract visual foreground that is universal, and then feed clean invariant vision to a standard RL policy learner (\fig{frameworkSimple}).  As the visual observation varies little between training and testing, the RL policy can be simplified and focused, delivering far better results.

Our technical challenge is to deliver such  clean visuals with a completely unsupervised learning approach, without mannual annotations or access to environment internals.

Given videos of actions in a training environment, we first learn how to extract visual foreground with unsupervised keypoint detection followed by unsupervised visual attention to automatically generate a foreground mask per video frame.  We can then introduce artificial distractors and train a model  to reconstruct the clean foreground mask from noisy observations.  Only this learned model, not the keypoint or attention model, is needed during test to provide distraction-free visual input to the RL policy learner.

Our unsupervised Visual Attention and Invariance (VAI) method has several desirable properties.
\begin{enumerate}[leftmargin=*,itemsep=-3pt,topsep=2pt]
\item 
{\bf Unsupervised task-agnostic visual adaption training}.
Our foreground extraction 
only assumes little background change between adjacent video frames, requiring no manual annotations or knowledge of environment internals (e.g. get samples with altered textures).
It does not depend on the task, policy learning, or task-specific rewards associated with RL.  That is, for different tasks in the same environment, we only need to collect one set of visual observations and train one visual adapter, which gets us a huge saving in  real-world robotic applications.

\item{\bf Stable policy training, no test-time adaptation.}
By freeing RL from visual distractions, our policy learning is stable and fast without being subject to strong domain augmentations, and our policy deployment is immediate without test-time fine-tuning.

\item{\bf Clear interpretation and modularization.}
We extract keypoints from videos to identify foreground, based on which attentional masks can be formed.  This unsupervised foreground parsing allows us to anticipate visual distractions and train a model to restore clean foregrounds.  
Compared to existing methods that work on intermediate features, our method has clear assumptions at each step, which can be  visualized, analyzed, and improved.

\end{enumerate}

We conduct experiments on two challenging benchmarks with diverse simulation environments: 
DeepMind Control suite 
\cite{tassa2018deepmind,hansen2020self} and our DrawerWorld robotic manipulation tasks with texture distortions and background distractions during deployment. 
Our VAI significantly outperforms the state-of-the-art, gaining 15$\sim$49\% (61$\sim$229\%) more cumulative rewards per episode on DeepMind Control (our DrawerWorld Manipulation) benchmarks.  

To summarize, we make the following contributions.
\begin{enumerate}[leftmargin=*,itemsep=-3pt,topsep=2pt]
\item 
We propose a novel domain generalization approach for vision-based RL: Instead of learning a universal policy for varying visual domains, we decouple vision and action, learning to extract universal visual foreground while keeping the RL policy learning intact.

\item 
We propose a fully unsupervised, task-agnostic visual adaptation method
that removes unseen distractions and restores clean foreground visuals.
Without manual annotations, strong domain augmentations, or test-time adaptation, our policy training is stable and fast, and our policy deployment is immediate without any latency.

\item 
We build unsupervised keypoint detection based on KeyNet \cite{jakab2018unsupervised} and Transporter \cite{kulkarni2019unsupervised}.  We develop a novel unsupervised visual attention module with causal inference for counterfactual removal.  We achieve visual invariance by unsupervised distraction adaptation based on foreground extraction.  Each step is modularized and has clear interpretations and visualizations.

\item
We propose DrawerWorld, a pixel-based robotic manipulation benchmark, to test the adaptation capability of vision-based RL to various realistic textures.

\item
Our results demonstrate that it is not only possible to learn domain-invariant vision from videos without supervision, but freeing RL from visual distractions also leads to better policies, setting new SOTA by a large margin.
\end{enumerate}


\section{Related Works}

\noindent\textbf{Unsupervised Learning} has made much progress in natural language processing, computer vision, and RL.  It aims to learn a feature transferable to downstream tasks  \cite{doersch2015unsupervised,noroozi2016unsupervised,larsson2017colorization,devlin2018bert,wu2018unsupervised, he2020momentum,chen2020improved,chen2020simple,wang2020unsupervised}.  
In RL,
UNREAL \cite{jaderberg2016reinforcement} proposes unsupervised reinforcement and auxiliary learning to improve learning efficiency of model-free RL algorithms, by maximizing pseudo-reward functions;
CPC \cite{oord2018representation} learns representations for RL in 3D environments by predicting the future in the latent space with autoregressive models;
CURL \cite{srinivas2020curl} extracts high-level features from raw pixels using contrastive learning and performs off-policy control on extracted features to improve data-efficiency on pixel-based RL.

\noindent\textbf{Domain Adaptation} incorporates an adaptation module to align the feature distribution from the source domain and the target domain without paired data \cite{patel2015visual,long2015learning,rosenfeld2018incremental,mallya2018piggyback,zhu2017unpaired,ghifary2016deep}. There are various approaches to
this, from using supervised data \cite{zhuang2015supervised, long2015learning, rosenfeld2018incremental}, to assumed correspondences \cite{gupta2017learning}, to unsupervised approaches \cite{ammar2015unsupervised, tzeng2017adversarial, zhu2017unpaired}.

\noindent\textbf{Multi-domain Learning} learns representations for multiple domains known a prior \cite{joshi2012multi,nam2016learning,rebuffi2017learning,wang2019towards}.  A combination of shared and domain-specific parameters are adopted.   It is also feasible to simply learn multiple visual domains with residual domain adapters \cite{rebuffi2017learning,wang2019towards}.

\noindent
Our work is different from these works, since we do not have prior knowledge of test data distributions and the model needs to generalize to unknown test environments.

\noindent\textbf{Robustness to Distribution Shifts} studies the effect of corruptions, perturbations, out-of-distribution examples, and real-world distribution shifts \cite{nettleton2010study,madry2017towards,hendrycks2019using,morimoto2005robust,pinto2017robust,heinrich2016deep}.  Recent deep RL approaches model such uncertainties explicitly.  

\noindent
\cite{heess2015memory} uses recurrent neural networks for direct adaptive control and determines dynamic model parameters on-the-fly. UP-OSI \cite{yu2017preparing} applies indirect adaptive control for online parameter identification.  EPOpt \cite{rajeswaran2016epopt} uses simulated source domains and  adversarial training to learn policies that are robust and generalizable to a range of possible target domains.  PAD \cite{hansen2020self} uses self-supervision to continue policy training during deployment without any rewards, achieving SOTA in several environments. SODA \cite{hansen2020generalization}, a concurrent work to ours,  alternates strong augmentations associated with self-supervised learning and weak augmentations associated with RL for obtaining both generalizability and stability.

\noindent
Instead of demanding a universal {\it policy} that is invariant to distribution shifts or transferable to novel environments, we achieve generalizability by demanding universal {\it visuals} that can be fed into the subsequent RL policy learner,  freeing it from visual distractions and making it more effective.
\def\figFramework#1{
\begin{figure*}[#1]
\centering
\includegraphics[width=0.99\textwidth]{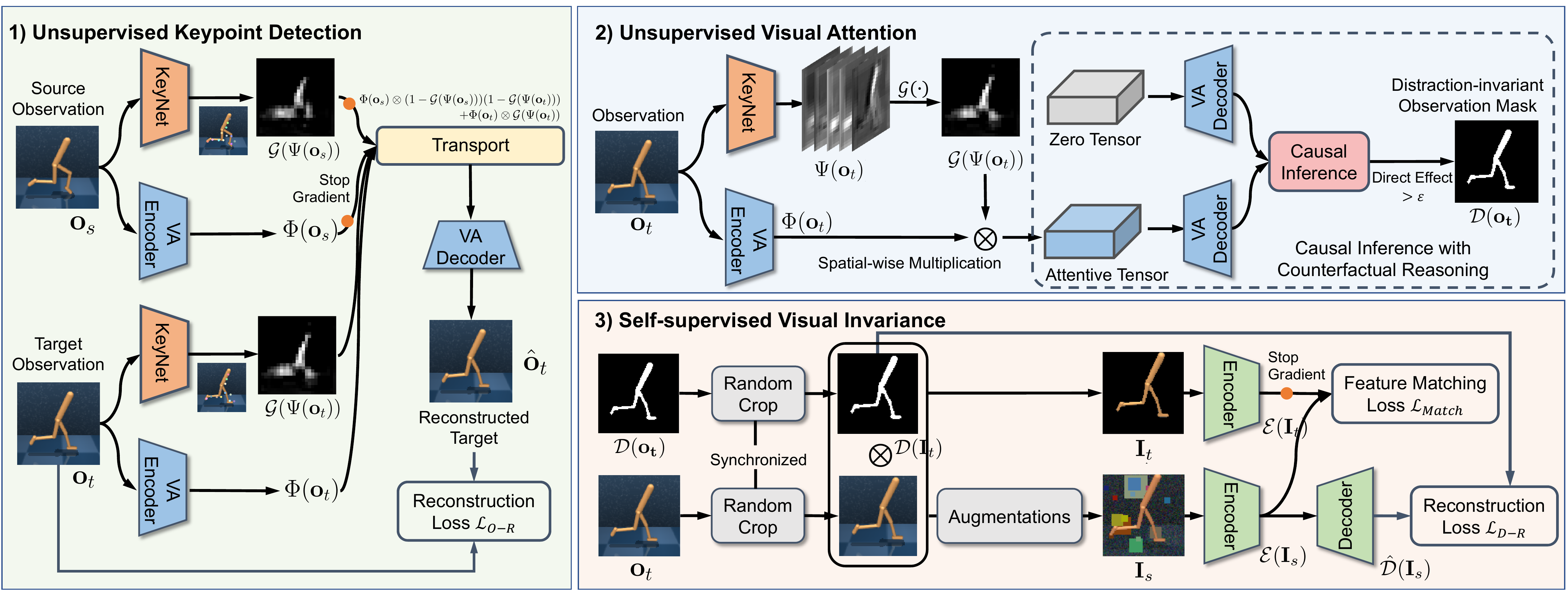}\vspace{-1pt}
\caption{Technical implementation of our   three components. 
{\bf 1) Unsupervised keypoint detection}:  
We build unsupervised keypoint detection and visual feature extraction based on KeyNet \cite{jakab2018unsupervised} and Transporter \cite{kulkarni2019unsupervised}.  The goal is to reconstruct the target frame from the target foreground appearance and the source-transported background appearance, capturing a moving foreground on a relatively still background.
{\bf 2) Unsupervised visual attention}:
We remove the model bias in the foreground mask derived from detected keypoints with novel causal inference for counterfactual removal.
{\bf 3) Self-supervised visual invariance}: We train a model to restore an invariant foreground visual image by adding artificial  distractors to extracted foreground and perform self-supervised distraction removal.
}
\label{fig:framework}
\end{figure*}
}

\def\figTracking#1{
\begin{figure}[#1]
\centering
\includegraphics[width=0.48\textwidth]{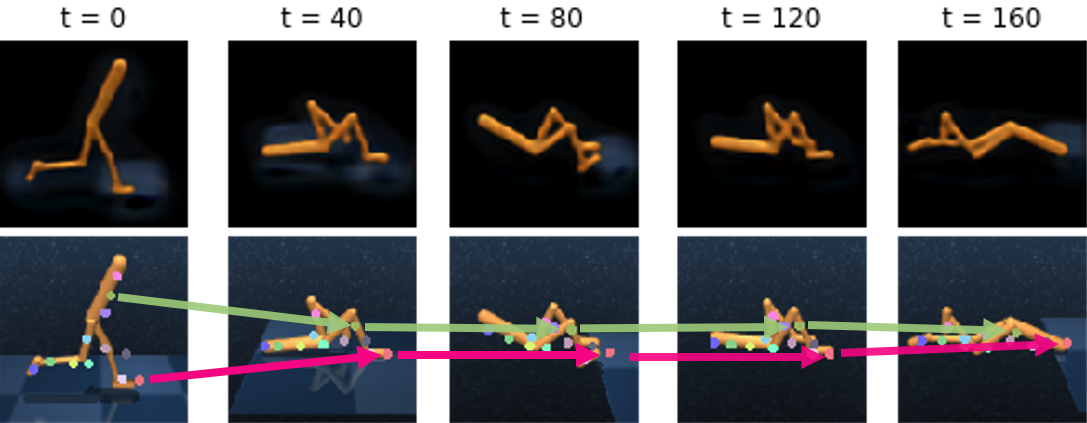}\vspace{-2pt}
\caption{
Our VAI foreground reconstruction (Row 1) provides clearer and more robust foreground visual information than detecting keypoints across image frames using Transporter (Row 2).
Due to occlusion, symmetry, and lacking visual distinctions,
it is often impossible to track keypoints consistently across frames.  That is, keypoint locations alone are not suitable as an invariant visual representation.
}\vspace{-2pt}
\label{fig:tracking}
\end{figure}
}

\def\figTracking#1{
\begin{figure}[#1]
\centering
\includegraphics[width=0.48\textwidth]{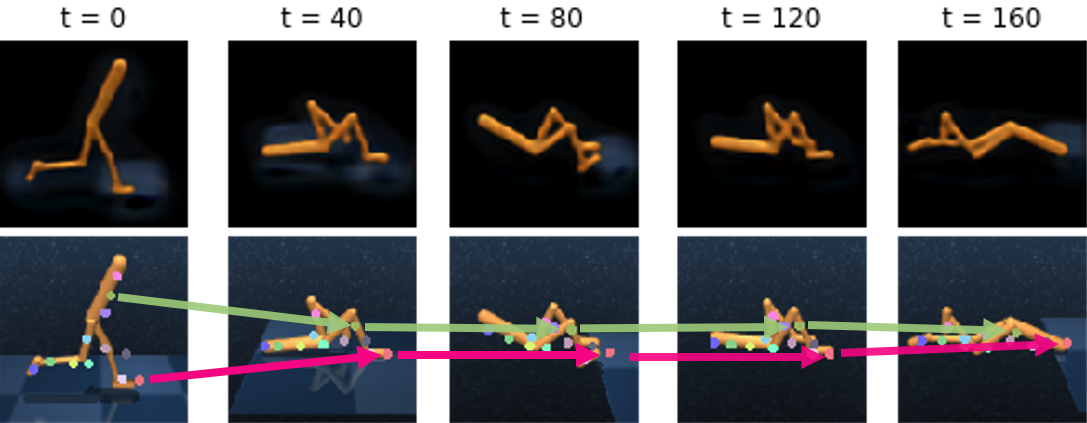}\vspace{-2pt}
\caption{
Our VAI foreground reconstruction (Row 1) provides clearer and more robust foreground visual information than detecting keypoints across image frames using Transporter (Row 2).
Due to occlusion, symmetry, and lacking visual distinctions,
it is often impossible to track keypoints consistently across frames.  That is, keypoint locations alone are not suitable as an invariant visual representation.
}\vspace{-2pt}
\label{fig:tracking}
\end{figure}
}

\def\figTDE#1{
\begin{figure}[#1]
\centering
\includegraphics[width=0.48\textwidth]{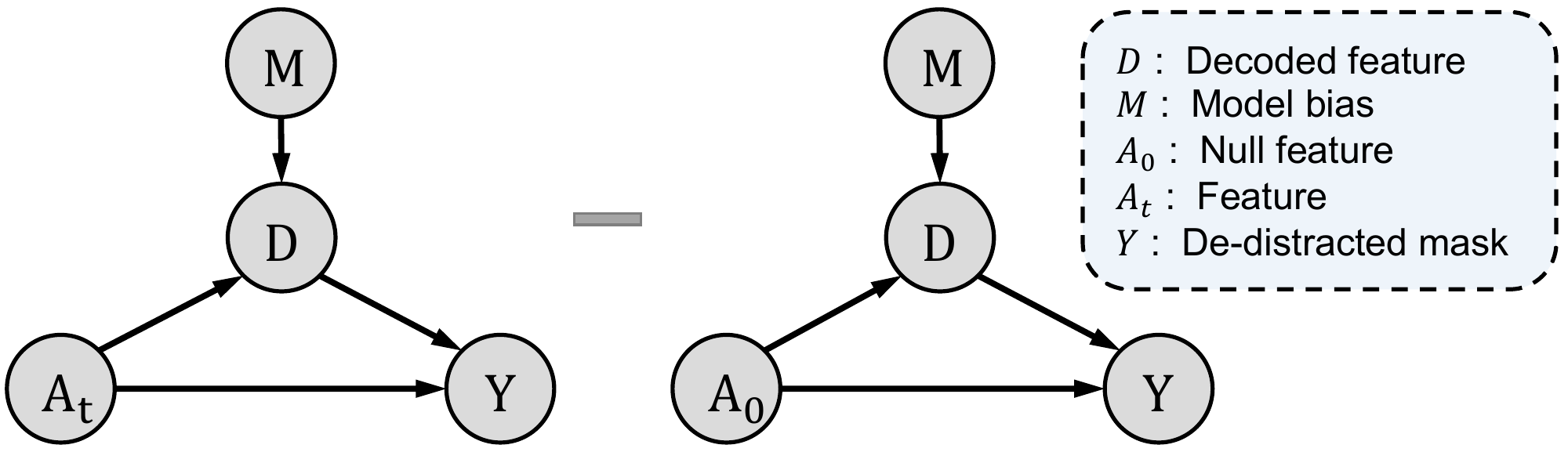}\vspace{-2pt}
\caption{Causal graph of casual inference with counterfactual reasoning for our foreground mask extraction.  The Controlled Direct Effect (CDE) is measured by the contrast between two outcomes: the counterfactual outcome given the visual feature $A_t$ and that given the null feature $A_0$.  
}
\label{fig:tde}
\vspace{-2pt}
\end{figure}
}

\def\imrow#1#2#3#4#5#6#7{
\includegraphics[width=0.166\linewidth]{figures/CDE/#1.png} & 
\includegraphics[width=0.166\linewidth]{figures/CDE/#2.png} & 
\includegraphics[width=0.166\linewidth]{figures/CDE/#3.png} & 
\includegraphics[width=0.166\linewidth]{figures/CDE/#4.png} & 
\includegraphics[width=0.166\linewidth]{figures/CDE/#5.png} & 
\includegraphics[width=0.166\linewidth]{figures/CDE/#6.png}\\[#7pt]
}

\def\figCDEVis#1{
\begin{figure}[#1] \centering
\setlength{\tabcolsep}{0pt}
\begin{tabular}{ccc|ccc}
\imrow
{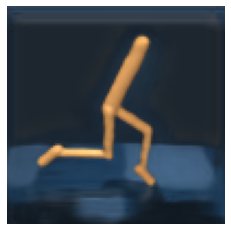}
{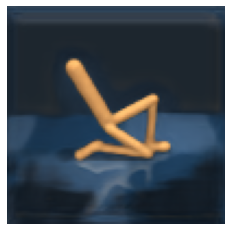}
{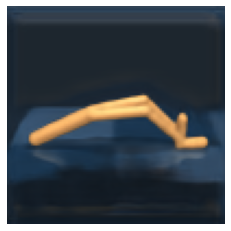}
{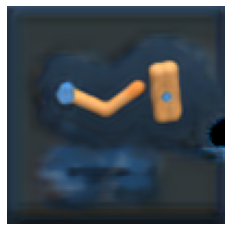}
{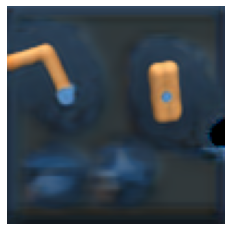}
{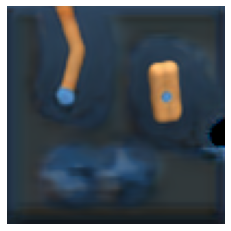}
{-5}
\imrow
{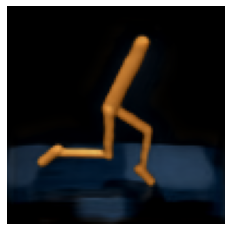}
{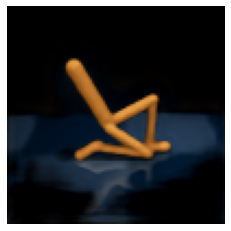}
{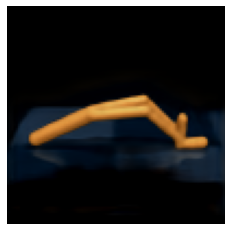}
{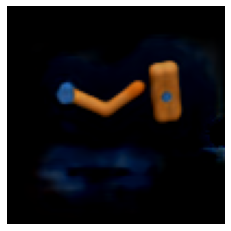}
{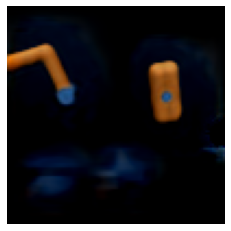}
{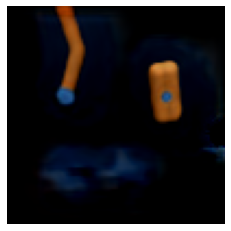}
{-4}
\multicolumn{3}{c}{walker walk} \vline&
\multicolumn{3}{c}{finger spin}\\[-2pt]
\end{tabular}
\caption{Foreground reconstructions with causal inference are cleaner (Row 2) than those without (Row 1). 
\label{fig:CDEVisual}
}
\end{figure}
}

\def\imsix#1#2#3#4#5#6#7#8{
\includegraphics[width=0.166\linewidth]{figures/KeyPoints/#1#8.png} & 
\includegraphics[width=0.166\linewidth]{figures/KeyPoints/#2#8.png} & 
\includegraphics[width=0.166\linewidth]{figures/KeyPoints/#3#8.png} & 
\includegraphics[width=0.166\linewidth]{figures/KeyPoints/#4#8.png} & 
\includegraphics[width=0.166\linewidth]{figures/KeyPoints/#5#8.png} & 
\includegraphics[width=0.166\linewidth]{figures/KeyPoints/#6#8.png}\\[#7pt]
}

\def\figKeypointVis#1{
\begin{figure}[#1]\centering
\setlength{\tabcolsep}{0pt}
\begin{tabular}{cc|cc|cc}
\imsix
{cartpole1}
{cartpole2}
{walker1}
{walker4}
{robot}
{robot3}
{-5}{-keypoint}
\imsix
{cartpole1}
{cartpole2}
{walker1}
{walker4}
{robot}
{robot3}
{-4}{-mask}
\multicolumn{2}{c}{cartpole} \vline&
\multicolumn{2}{c}{walker} \vline&
\multicolumn{2}{c}{open drawer} \\[-2pt]
\end{tabular}
\caption{Our detected keypoints (Row 1) and generated foreground masks (Row 2) from DeepMind control and DrawerWorld benchmarks.  Note that they could cover multiple moving objects in the foreground. 
\label{fig:keypointVis}
}
\end{figure}
}

\def\imrowIMG#1#2#3#4#5#6#7#8#9{
\includegraphics[width=#9\linewidth, height=#9\linewidth]{figures/DeepMindTests/#1.png} & 
\includegraphics[width=#9\linewidth, height=#9\linewidth]{figures/DeepMindTests/#2.png} & 
\includegraphics[width=#9\linewidth, height=#9\linewidth]{figures/DeepMindTests/#3.png} & 
\includegraphics[width=#9\linewidth, height=#9\linewidth]{figures/DeepMindTests/#4.png} & 
\includegraphics[width=#9\linewidth, height=#9\linewidth]{figures/DeepMindTests/#5.png} & 
\includegraphics[width=#9\linewidth, height=#9\linewidth]{figures/DeepMindTests/#6.png} & 
\includegraphics[width=#9\linewidth, height=#9\linewidth]{figures/DeepMindTests/#7.png}\\[#8pt]
}

\def\figTestEnvVis#1{
\begin{figure*}[#1] \centering
\setlength{\tabcolsep}{0pt}
\begin{tabular}{c|cc|cc|cc}
\imrowIMG
{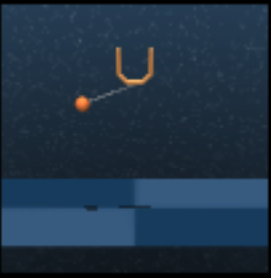}
{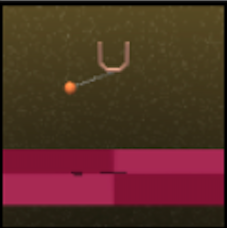}
{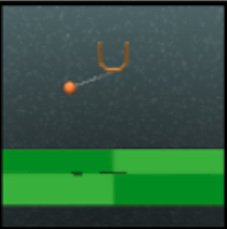}
{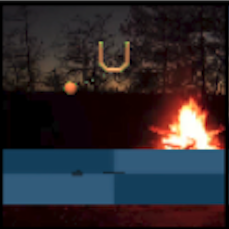}
{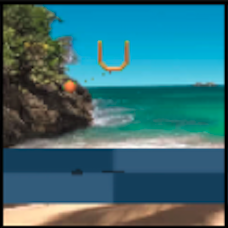}
{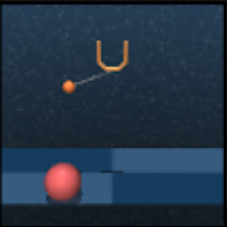}
{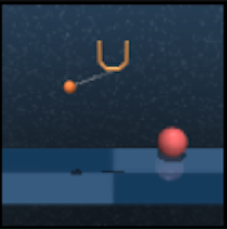}
{-2}
{0.14}
\multicolumn{1}{c}{training} \vline&
\multicolumn{2}{c}{randomized colors} \vline&
\multicolumn{2}{c}{video backgrounds} \vline&
\multicolumn{2}{c}{distractions}\\[-7pt]
\end{tabular}
\caption{Visualization results of testing environments in DeepMind Control benchmark \cite{hansen2020self, tassa2018deepmind}. The testing environment changes include randomized colors, video backgrounds, and background distractions.
\label{fig:TestEnvVis}
}
\end{figure*}
}

\section{Unsupervised Visual Attention \& Invariance}

\figFramework{!t}

Our goal is to extract universal visual foreground and then feed clean invariant vision to an RL policy learner (\fig{frameworkSimple}).  Our technical challenge is to deliver such clean visuals with a completely unsupervised learning approach, without mannual annotations or access to environment internals.

Our VAI method has three components: 
Unsupervised keypoint detection, 
unsupervised visual attention,
and self-supervised visual invariance.
The first two are only used during training to extract foregrounds from videos without supervision, whereas only the last trained model is deployed to automatically remove distractors from a test video.

\subsection{Unsupervised Keypoint Detection}

We assume that training videos contain  moving foregrounds against a relatively still background.  Our idea for unsupervised foreground extraction is the following:  Given two such source and target frames, we can learn to predict keypoints and visual features from each image so that foreground features and target keypoints can be used to reconstruct the target frame, without requiring manual annotations. 

For a particular image pair, the moving foreground may have a still part ({\it upper body}), or the background may have a moving part ({\it flickering flames}) .  However, when the keypoint predictor and the visual feature extractor have to work consistently across all the videos in the same environment, they would have to focus on the entire moving foreground and disregard the random minor background motion.

Let ${\bf{o}}_s, {\bf{o}}_t \!\in\! \mathbb{R}^{C\!\times H \!\times\! W}$ denote the source and target frames sampled from a trajectory, where $C$, $H$, and $W$ are the channel dimension, image height and width respectively.
Let $\Phi(\cdot)$ denote the visual feature extractor.
Let ${\Psi(\cdot)}$ denote the keypoint network that predicts $K$ keypoints in terms of 2D spatial locations $\{\mu_k\}$.  We render each keypoint as a smaller $H'\times W'$ Gaussian heatmap with fixed variance  $\sigma^2$, and derive a foreground mask by taking the max of all of them:
\begin{equation}
\mathcal{G}(\mu;x) = \max_{k\in\{1, 2, ..., K\}} \exp\left(-\frac{\|x - \mu_{k}\|^2}{2\sigma^2}\right).
\end{equation}

We follow KeyNet \cite{jakab2018unsupervised, kulkarni2019unsupervised} to reconstruct the target observation ${\bf{o}}_t$ from $K$ landmarks ${\Psi({\bf{o}}_s)},{\Psi({\bf{o}}_t)}$.
We follow \cite{kulkarni2019unsupervised} to transport the source background appearance to the target frame by putting the source feature at common background areas and the target feature at the target keypoints:
\begin{align}
\hat{\Phi}({\bf{o}}_t, {\bf{o}}_s) =&  
\Phi({\bf{o}}_s) \otimes (1 - \mathcal{G}(\Psi({\bf{o}}_s)))(1 - \mathcal{G}(\Psi({\bf{o}}_t)))  
\nonumber\\
+&
\Phi({\bf{o}}_t) \otimes 
\mathcal{G}(\Psi({\bf{o}}_t))
\end{align}
where $\otimes$ denotes location-wise multiplication applied to each channel.
A visual attention (VA) decoder outputs a reconstruction $\hat{{\bf{o}}}_t$ of target frame ${\bf{o}}_t$ from the transported feature $\hat{\Phi}({\bf{o}}_t, {\bf{o}}_s)$.
Minimizing the reconstruction loss below optimizes the KeyNet and the visual attention encoder/decoder end-to-end:
\begin{equation}
\mathcal{L}_{\text{O-R}}({\bf{o}}_t,\hat{\bf{o}}_t)=\|{\bf{o}}_t-\hat{\bf{o}}_t\|_2^2.
\end{equation}

Note that the original Transporter only focuses on changes between frames in the same episode, 
whereas we also sample frames from different episodes 50\% of the time in reacher environment which has a fixed target throughout each episode, so that our keypoints will be able to capture the target and spread over the entire moving foreground.

\subsection{Unsupervised Spatial Attention}

Now we already have an unsupervisedly learned keypoint detector.  We first explain why we do not use keypoints for control and instead derive a visual foreground mask. We then describe our novel causal inference formulation for obtaining a foreground mask without model bias.

\figTracking{!t}

Transporter \cite{kulkarni2019unsupervised} successfully makes use of keypoints for RL in Atari ALE \cite{bellemare2013arcade} and Manipulator \cite{tassa2018deepmind}. Keypoints are geometrical extraction without visual appearance distractions that they could be potentially used to minimize differences between training and testing environments.

However, there are three major issues with keypoints in practice. {\bf 1)} It is often hard to track keypoints consistently across frames; even for humans, whether a keypoint is on the left or right foot is unclear in \fig{tracking}. This implies that using predicted keypoints for control directly would be brittle even in clean images.

{\bf 2)} While keypoints along with image features and LSTM could work on relatively complicated tasks  \cite{kulkarni2019unsupervised}, they add substantial model complexity and computational costs.  
{\bf 3)} While keypoints themselves are free of visual distractions,  
their extractor ({\it KeyNet}) is only trained for the training environment, with no guarantee for robustness against domain shifts.

\figCDEVis{!t}
\figTDE{!t}

We thus propose to generate a foreground mask $\mathcal{G}({\Psi({\bf{o}}_t)})$ from the ({un-ordered}) collection of predicted keypoints instead.  
We enhance the visual feature in the foreground: $\mathcal{G}({\Psi({\bf{o}}_t)}) \otimes \Phi({\bf{o}}_t)$ and pass it to the VA decoder to reconstruct a cleaner image $\hat{\bf{o}}_t$ (\fig{CDEVisual} Row 1).  However, it is blurry with common background remnants captured in the bias terms of the decoder.  The bias terms are essential for proper reconstruction and  cannot be simply set to zero. 

We apply causal inference with counterfactual reasoning \cite{pearl2013direct,richiardi2013mediation,pearl2018book,greenland1999confounding,tang2020long} to remove the model bias (Fig. \ref{fig:tde}).
 Intuitively, the predicted foreground mask $Y$ has a direct cause from the visual feature $A$, and an indirect cause from the model bias $M$ through the decoder $D$.  To pursue the direct causal effect, we perform counterfactual reasoning
known as {\it Controlled Direct Effect} (CDE) \cite{pearl2013direct,greenland1999confounding}, which
contrasts the counterfactual outcomes (marked by $do(\cdot)$)
between visual feature $A_t$ and null visual feature $A_0$ (set to the zero tensor):
\begin{equation}
\!\!\text{CDE}(Y)\! =\! [Y|do(A_t), do(M)] \!-\! [Y|do(A_0), do(M)].
\end{equation}
\noindent
We further threshold it to obtain the foreground mask $\mathcal{D}({\bf{o}_t})$:
\begin{align}
\mathcal{D}({\bf{o}_t})=\begin{cases}
               0, \text{CDE}(Y) < \epsilon\\
              1, \text{CDE}(Y) \geq \epsilon
            \end{cases}.
\end{align}
Fig.~\ref{fig:keypointVis} shows our detected keypoints and generated masks: {\bf 1)} While the keypoints may be sparse and imprecise, the foreground mask is clean and complete; 
{\bf 2)} Our unsupervisedly learned keypoints do not correspond to semantic joints of articulation, e.g., for the grasper opening a drawer, there are keypoints on both the grasper and the drawer, and our derived foreground mask contains both moving objects.

\figKeypointVis{t}

\figTestEnvVis{t}

\subsection{Self-supervised Visual Invariance}
\label{section:visual_invariance}

Our spatial attention module outputs a foreground mask, after seeing samples in the training environment.  To make it adaptable to unknown test environments, we augment the clean foreground image with artificial distractors and train a model to reconstruct a mask to retrieve clean foreground observation.

Given image ${\bf o}_t$, we generate equally cropped clean target image $I_t$ and noisy source image $I_s$.
\begin{align} 
I_t & = \mathcal{T}_c({\bf{o}}_t) \otimes\mathcal{T}_c(\mathcal{D}({\bf{o}}_t)) \\
I_s & =\mathcal{T}_f(I_t) + \mathcal{T}_b(\mathcal{T}_c({\bf{o}}_t) \otimes(1 - \mathcal{D}(I_t)))
\end{align}
where $\mathcal{D}(I_t) = \mathcal{T}_c(\mathcal{D}({\bf{o}}_t))$, $\mathcal{T}_c$ denotes synchronized random crop, $\mathcal{T}_f$ adds possible foreground changes such as color jitter and random brightness change, whereas $\mathcal{T}_b$ adds a set of possible background changes such as random colored boxes to the background.
We learn a convolutional encoder/decoder pair to reconstruct the clean foreground mask from noisy $I_s$, so that they could focus more on the foreground and ignore background distractors.  We impose a feature matching loss at output of encoder $\mathcal{E}$ and an image reconstruction loss at output of decoder $\hat{\mathcal{D}}$:
\begin{equation}
\mathcal{L}_{\text{total}} =
\|\hat{\mathcal{D}}({\bf{I}}_s) - \mathcal{D}({\bf{I}}_t)\|_2 ^2 + 
\lambda \cdot \|\mathcal{E}({\bf{I}}_s)-\mathcal{E}({\bf{I}}_t)\|_2 ^2
\end{equation}
where $\mathcal{D}({\bf{I}}_t)$ is simply the cropped version of 
$\mathcal{D}({\bf{o}}_t)$.  During RL training and deployment, for any frame $I$, we feed $I\otimes \hat{\mathcal{D}}(\mathcal{E}(I))$ to the learned RL policy.

What augmentations to use has a big impact on generalization.
We propose four additional strong background augmentations on $\mathcal{T}_b$.
{\bf 1)} The background could randomly assume the training image background, a random color,  or the mean foreground color with small perturbations. 
{\bf 2)} Gaussian pixel-wise noise and random boxes are added.  MultiColorOut, an extension to Cutout-color \cite{laskin_lee2020rad}, adds multiple boxes of random sizes, colors, and positions.
{\bf 3)} Darkened foreground copies are added to the background areas where the foreground mask values are 0, to simulate distractors that look similar to the foreground.
{\bf 4)} We follow \cite{hansen2020generalization} to randomly select images in the Places dataset  \cite{zhou2017places} as background images for augmentation. For fair comparisons, we list our results with and without this option.
With such generic augmentations, our model is able to perform well on realistic textures and unknown testing environments even though it has not encountered them during training.

{\bf RL policy training with weak augmentations.}  Our visual invariance model outputs a clean foreground image with background distractors suppressed.  The RL policy learner still needs to handle foreground variations in unknown test environments.  We train our RL policy with weak foreground augmentations to make it robust to noise and distortions.  We add the usual Gaussian random noise and use only a simple MultiColorOut to simulate the inclusion of backgrounds and missing foreground parts.  Empirically we find that such weak augmentations do not affect the RL training stability.

\def\figDrawer#1{
\begin{figure}[#1]
\centering
\includegraphics[width=0.8\linewidth]{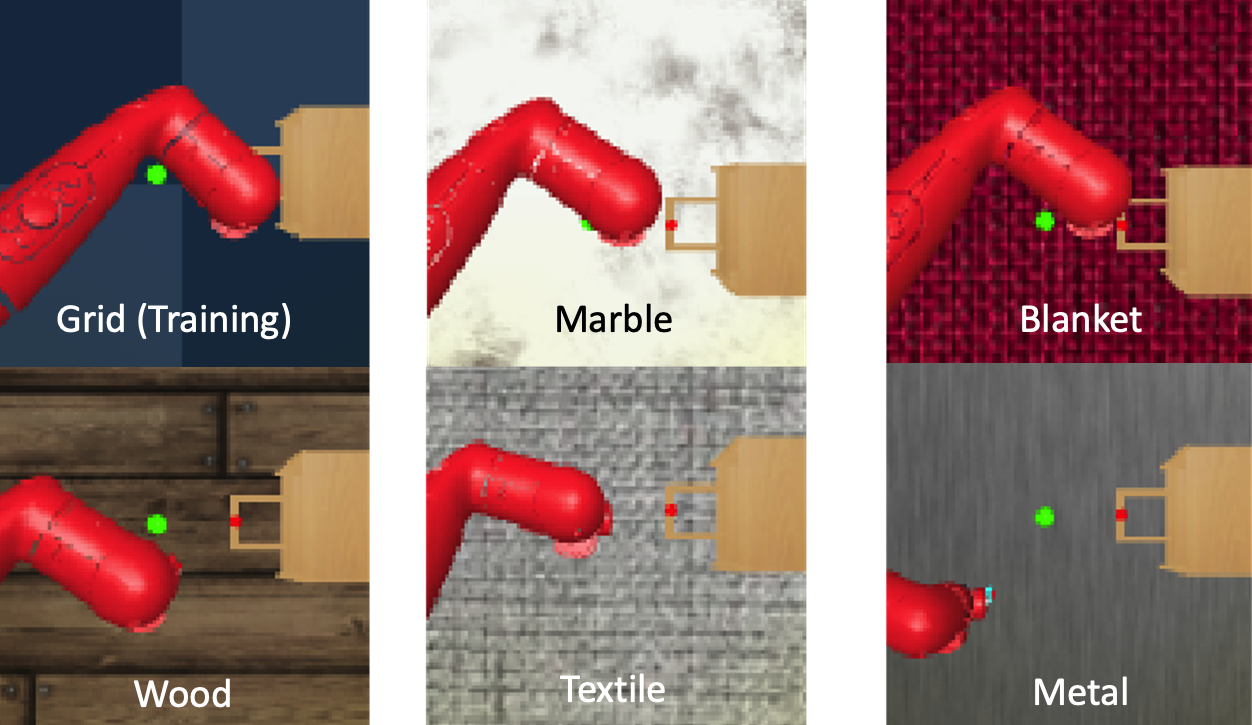}
\vspace{-2pt}
\caption{DrawerWorld environments. Grid is the texture used in training. Other 5 evaluation textures are from realistic photos, which makes the task challenging.}
\label{fig:drawer}
\end{figure}
}

\newcommand{\upa}[1]{{\scriptsize ($\uparrow$#1\%)}}
\newcommand{\doa}[1]{{\scriptsize ($\downarrow$#1\%)}}
\newcommand{\deltacomp}[2]{{\begin{tabular}[x]{@{}l@{}}

\ifnum #1 > #2
{\color{green(pigment)} \bf +\the \numexpr #1-#2\relax}
\else
{\color{red} \bf \the \numexpr #1-#2\relax}
\fi \\[-4pt]
\ifnum #2 = 0
{\scriptsize{($\uparrow\infty$\%)}}
\else
\ifnum #1 > #2
{\scriptsize{($\uparrow$ \the \numexpr (#1-#2)*100/(#2)\relax \%)}}
\else
{\scriptsize{($\downarrow$ \the \numexpr (#2-#1)*100/(#2)\relax \%)}}
\fi
\fi
\vspace{-2pt}
\end{tabular}
}}

\newcommand{\score}[2]{{\begin{tabular}[x]{@{}c@{}}
#1 \\[-4pt]
\scriptsize{\textpm#2}
\vspace{-2pt}
\end{tabular}
}}

\def\tabColor#1{
\begin{table}[#1]
    \tablestyle{1.6pt}{1.1}
    \centering
    \begin{tabular}{l||cccc|ccl}
    \shline
    Random colors & SAC  & DR     & PAD    & SODA+P  & VAI & VAI+P & $\Delta$ \\
    \shline
    Walker, walk       & \score{414}{74}  & \score{594}{104} & \score{468}{47} & \score{692}{68} & \score{819}{11} & \textbf{\score{918}{6}} & \deltacomp{918}{692} \\
    Walker, stand      & \score{719}{74}  & \score{715}{96}  & \score{797}{46} & \score{893}{12} & \textbf{\score{964}{2}} & \textbf{\score{968}{3}} & \deltacomp{968}{893} \\
    Cartpole, swingup  & \score{592}{50}  & \score{647}{48}  & \score{630}{63} & \score{805}{28} & \textbf{\score{830}{10}} & \score{819}{6} & \deltacomp{819}{805} \\
    Cartpole, balance  & \score{857}{60}  & \score{867}{37}  & \score{848}{29} & - & \textbf{\score{990}{4}} & \score{957}{9} & \deltacomp{990}{848} \\
    Ball in cup, catch & \score{411}{183} & \score{470}{252} & \score{563}{50} & \score{949}{19} & \score{886}{33} & \textbf{\score{960}{8}} & \deltacomp{960}{949} \\
    Finger, spin       & \score{626}{163} & \score{465}{314} & \score{803}{72} & \score{793}{128} & \score{932}{3} & \textbf{\score{968}{6}} & \deltacomp{968}{803} \\
    Finger, turn\_easy  & \score{270}{43}  & \score{167}{26}  & \score{304}{46} & - & \textbf{\score{445}{36}} & \textbf{\score{455}{48}} & \deltacomp{455}{304} \\
    Cheetah, run       & \score{154}{41}  & \score{145}{29}  & \score{159}{28} & - & \textbf{\score{337}{1}} & \textbf{\score{334}{2}} & \deltacomp{337}{159} \\
    Reacher, easy      & \score{163}{45}  & \score{105}{37}  & \score{214}{44} & - & \textbf{\score{934}{22}} & \textbf{\score{936}{19}} & \deltacomp{936}{214} \\
    \hline
    \it average       &\it 467  &\it 464  &\it 531 & - & \it{793} & \textbf{\textit{812}} &\textit{\deltacomp{812}{531}} \\
    \shline
    \end{tabular}\vspace{-2pt}
    \caption{
    VAI outperforms existing methods on DeepMind
    randomized color tests by a large margin 
    \textit{without} using the external Places dataset; it is even better than SODA+P, which uses Places as a part of the training set.
    Soft Actor-Critic (SAC) \cite{haarnoja2018soft,mnih2013playing} is used as a base algorithm for DR (domain randomization), PAD \cite{hansen2020self}, SODA \cite{hansen2020generalization}, and our VAI. SODA+P and VAI+P use Places \cite{zhou2017places} as overlay or adapter augmentation.
    The results of SAC and DR are copied from PAD \cite{hansen2020self}.  Listed are the mean and std of cumulative rewards across 10 random seeds and 100 random episode initializations per seed.  The absolute and relative improvement of VAI over SOTA method are listed in the $\Delta$ column. }
    \label{table:random_color} 
\end{table}
}

\def\tabVideo#1{
\begin{table}[#1]
    \tablestyle{0.2pt}{1.1}
    \centering
    \begin{tabular}{l||ccccc|ccl}
    \shline
    Video background & SAC  & DR     & PAD    & SODA & SODA+P & VAI & VAI+P & $\Delta$ \\
    \shline
    Walker, walk       & \score{616}{80}  & \score{655}{55} & \score{717}{79} & \score{635}{48} & \score{768}{38} & \score{870}{21} & \textbf{\score{917}{8}} & \deltacomp{917}{768} \\
    Walker, stand      & \score{899}{53}  & \score{869}{60}  & \score{935}{20} & \score{903}{56} & \score{955}{13} & \textbf{\score{966}{4}} & \textbf{\score{968}{2}} & \deltacomp{968}{955} \\
    Cartpole, swingup  & \score{375}{90}  & \score{485}{67}  & \score{521}{76} & \score{474}{143} & \textbf{\score{758}{62}} &  \score{624}{146} & \textbf{\score{761}{127}} & \deltacomp{761}{758} \\
    Cartpole, balance  & \score{693}{109}  & \score{766}{92}  & \score{687}{58} & - & - &  \textbf{\score{869}{189}} & \score{847}{205} & \deltacomp{869}{687} \\
    Ball in cup, catch & \score{393}{175} & \score{271}{189} & \score{436}{55} & \score{539}{111} & \textbf{\score{875}{56}} &  \score{790}{249} & \score{846}{229} & \deltacomp{846}{875} \\
    Finger, spin       & \score{447}{102} & \score{338}{207} & \score{691}{80} & \score{363}{185} & \score{695}{97} &  \score{569}{366} & \textbf{\score{953}{28}} & \deltacomp{953}{695} \\
    Finger, turn\_easy  & \score{355}{108}  & \score{223}{91}  & \score{362}{101} & - & - &  \score{419}{50} & \textbf{\score{442}{33}} & \deltacomp{442}{362} \\
    Cheetah, run       & \score{194}{30}  & \score{150}{34}  & \score{206}{34} & - & - &  \textbf{\score{322}{35}} & \textbf{\score{325}{31}} & \deltacomp{325}{206} \\
    \hline
    \it average       &\it 497  &\it 470  &\it 569 & - & - & \it678 & \textbf{\textit{757}} & \textit{\deltacomp{757}{569}} \\
    \shline
    \end{tabular}\vspace{-2pt}
    \caption{VAI+P (VAI) outperforms PAD by more than 33\% (19\%) on challenging DeepMind video backgrounds.  Same settings and conventions as Table \ref{table:random_color}.}
    \label{table:video} 
\end{table}
}

\def\tabDistractor#1{
\begin{table}[#1]
    \tablestyle{6.2pt}{1.1}
    \centering
    \begin{tabular}{l||ccc|cl}
    \shline
    Distracting objects & SAC & DR & PAD & VAI & {$\Delta$} \\
    \shline
    Cartpole, swingup  & \score{815}{60}  & \score{809}{24} & \score{771}{64} & \textbf{\score{891}{0}} & \deltacomp{891}{771} \\
    Cartpole, balance  & \score{969}{20}  & \score{938}{35} & \score{960}{29} & \textbf{\score{993}{0}} & \deltacomp{993}{969} \\
    Ball in cup, catch & \score{177}{111}  & \score{331}{189} & \score{545}{173} & \textbf{\score{956}{4}} & \deltacomp{956}{545} \\
    Finger, spin & \score{652}{184}  & \score{564}{288} & \textbf{\score{867}{72}} & \score{805}{3} & \deltacomp{805}{867} \\
    Finger, turn\_easy & \score{302}{68}  & \score{165}{12} & \score{347}{48} & \textbf{\score{389}{18}} & \deltacomp{389}{347} \\
    \hline
    \it average     & \it 583  &\it 561  &\it 698 & \textbf{\textit{806}} & \textit{\deltacomp{806}{698}} \\
    \shline
    \end{tabular}\vspace{-2pt}
    \caption{VAI outperforms current SOTAs by more than 15\% on DeepMind Control distracting objects.  Although VAI performs worse than PAD on ``Finger, spin" task in terms of mean rewards, the reward variance is greatly reduced from 72 to 3 in std. Same settings and conventions as Table \ref{table:random_color}.
    }
    \label{table:distracting} 
\end{table}
}

\def\tabDrawer#1{
\begin{table}[#1]
    \tablestyle{2.3pt}{1.1}
    \centering
    \begin{tabular}{l||cccl|cccl}
    \shline
     \multirow{2}{*}{success \%} & \multicolumn{4}{c|}{DrawerOpen} & \multicolumn{4}{c}{DrawerClose} \\
     & SAC & PAD & VAI & $\Delta$ & SAC & PAD & VAI & $\Delta$ \\
    \shline
    Grid & \score{98}{2} & \score{84}{7} & \textbf{\score{100}{0}} & \deltacomp{100}{98} & \textbf{\score{100}{0}} & \score{95}{3} & \score{99}{1} & \deltacomp{99}{100} \\ \hline
    Black           & \score{95}{2} & \score{95}{3} & \textbf{\score{100}{1}} & \deltacomp{100}{95} & \score{75}{4}  & \score{64}{9} & \textbf{\score{100}{0}} & \deltacomp{100}{75} \\ 
    Blanket         & \score{28}{8} & \score{54}{6} & \textbf{\score{86}{6}}  & \deltacomp{86}{54} & \score{0}{0}   & \score{0}{0}  & \textbf{\score{85}{8}}  & \deltacomp{85}{0} \\ 
    Fabric          & \score{2}{1}  & \score{20}{6} & \textbf{\score{99}{1}}  & \deltacomp{99}{20} & \score{0}{0}   & \score{0}{0}  & \textbf{\score{74}{8}}  & \deltacomp{74}{0} \\ 
    Metal           & \score{35}{7} & \score{81}{3} & \textbf{\score{98}{2}}  & \deltacomp{98}{81}& \score{0}{0}   & \score{2}{2}  & \textbf{\score{98}{3}}  & \deltacomp{98}{2} \\ 
    Marble          & \score{3}{1}  & \score{3}{1}  & \textbf{\score{43}{7}}  & \deltacomp{43}{3} & \score{0}{0}   & \score{0}{0}  & \textbf{\score{49}{13}} & \deltacomp{49}{0} \\ 
    Wood            & \score{18}{5} & \score{39}{9} & \textbf{\score{94}{4}}  & \deltacomp{94}{39} & \score{0}{0}   & \score{12}{2} & \textbf{\score{70}{6}}  & \deltacomp{70}{12} \\ 
    \hline
    \it average     &\it 40 &\it 54 &\it \textbf{87}  &\textit{\deltacomp{87}{54}} &\it 25  &\it 25 &\it \textbf{82}  & \textit{\deltacomp{82}{25}} \\ 
    \shline
    \end{tabular}\vspace{-2pt}
    \caption{Our
    VAI consistently outperforms all the baselines in new texture environments, and on DrawerClose in particular, VAI succeeds 82\% vs. SAC/PAD's 25\%.  Grid is the training environment.
     Black means a completely dark background without texture. Other textures are shown in Fig. \ref{fig:drawer}.  DrawerClose is more challenging than DrawerOpen, as the drawer handle is concealed by the effector in DrawerClose, which would require the agent to infer the handle position from the position and the size of the effector. The success rate is the percentage of successful attempts out of 100 attempts to open or close a drawer.  The mean/std are collected  over 10 seeds.}
    \label{table:drawer}
\end{table}
}

\def\tabAblation#1{
\begin{table}[#1]
    \tablestyle{2.pt}{1.1}
    \centering
    \setlength{\tabcolsep}{7pt}
    \begin{tabular}{l||cccc}
    \shline
success rate (\%) & Grid & Wood & Metal          & Fabric \\ \shline
SAC & \score{98}{2} &\score{18}{5} & \score{35}{7} & \score{2}{1} \\ \hline
+ RL Augmentation &\textbf{\score{100}{1}} &\score{18}{5} & \score{41}{8} & \score{24}{5} \\ \hline
\begin{tabular}[c]{@{}l@{}} + Foreground Extraction\end{tabular}
 &\textbf{\score{100}{0}} & \score{18}{4}  & \score{13}{4} & \score{38}{4} \\ \hline
\begin{tabular}[c]{@{}l@{}} + Background Augmentation\end{tabular}
 &\textbf{\score{100}{0}} &\textbf{\score{94}{4}} & \textbf{\score{98}{2}} & \textbf{\score{99}{1}} \\
 \shline
    \end{tabular}\vspace{-2pt}
    \caption{Ablation studies for augmentation and foreground extraction on DrawOpen task.  From top to bottom rows,  components are added to the method cumulatively.
    Each method is trained in the grid environment and tested in new texture enviroments of wood, metal, and fabric.
    Success rates are collected over 500K steps.  
    Only the last method with all augmentations deliver consistent robustness.
    }
    \label{table:ablation}
    \vspace{-2pt}
\end{table}
}

\tabColor{!t}

\section{Experiments}
We experiment on two benchmarks, DeepMind and DrawerWorld, and perform ablation studies.
The DeepMind Control benchmark contains various background distractions \cite{tassa2020dmcontrol, hansen2020self} as in \fig{TestEnvVis}.
We propose a DrawerWorld Robotic Manipulation benchmark, based on MetaWorld \cite{yu2019meta}, in order to test a model's texture adaptability in manipulation tasks.

\subsection{DeepMind Control Benchmark} 
\noindent
{\bf Tasks.} There are  walking, standing, and reaching objects \cite{tassa2020dmcontrol}, all in 3D simulation.  Our agent receives pixel-based inputs instead of state-based inputs from the underlying dynamics unless otherwise stated. 

\noindent
{\bf Testing.} We follow PAD \cite{hansen2020self} and test our method under three types of environments: 1) randomized colors; 2) video backgrounds; and 3) distracting objects. For tasks with video background and distracting objects, we apply a moving average de-noising trick by subtracting a moving average of the past observations from the current observation and adding back the mean color of the moving average. We introduced a constant factor $\alpha$ multiplied to past moving average to tune the aggressiveness of the de-noising trick.

\noindent
{\bf Training.} For each scenario, we train agents without distractions and evaluate the model across 10 random seeds and 100 random environment initializations.  To get observation samples for training, we export 5000 transitions from the replay buffer for the training environment, which are collected with a random policy.  We use the same environment settings such as frame skip and data augmentation as in PAD to ensure fair comparisons between VAI, PAD, and others.

\tabVideo{!t}
\tabDistractor{!t}

\noindent
{\bf Randomized color results.}
Table \ref{table:random_color} shows that our VAI outperforms published SOTA on all the 9 tasks by up-to an astonishing 337\% margin in terms of mean cumulative rewards, without seeing samples in the test environment at any time.  
In contrast, DR is trained with color change to the environment (which requires knowing and changing the internals of the environment), which, to some extent, previews what the test environment would be.  Similarly, although PAD does not use any evaluation samples during training, it does use the samples at the test time to tune the encoder.  
Since VAI does not change model weights, it has no adaptation delay, better stability, and less compute (see more details in supplementary materials).  By suppressing distractions and feeding only the foreground image, the RL algorithm ideally sees the same input no matter what the environment is and is thus not influenced by background distractions or domain shifts in the test environment.

\noindent
{\bf Video background results.} 
Table \ref{table:video} shows that our VAI outperforms baselines in 7 out of 8 tasks in terms of mean cumulative rewards, often by a large margin.

\noindent
{\bf Distracting object results.}
Table \ref{table:distracting} shows that our VAI surpasses baselines on 4 out of 5 tasks in terms of cumulative rewards.  It not only obtains nearly full scores on ``Cartpole, balance" and ``Ball in cup, catch" tasks, but also greatly decreases the variance of results to a negligible level. 

\tabDrawer{!t}

\subsection{DrawerWorld Manipulation Benchmark}
\label{DrawerWorld}
\figDrawer{!t}

\noindent
{\bf A New Texture Benchmark for Manipulation.}
CNNs are sensitive to textures \cite{geirhos2018imagenet}.
We propose to evaluate a model's texture adaptability in manipulation tasks,
based on the MetaWorld \cite{yu2019meta} benchmark for meta RL and multi-task RL. 

In the original MetaWorld, the observations include 3D Cartesian positions of the robot, the object, and the goal positions collected with sensors on the object and the robot.  Accurate object positions and robot keypoints are hard to get by in real-world applications, we thus propose a variant of MetaWorld, {\it DrawerWorld}, with visual observations instead.  We focus on the variety of \textit{realistic textures} (Fig. \ref{fig:drawer}).

\noindent
{\bf Tasks.} There are DrawerOpen and DrawerClose tasks, where a Sawyer arm is manipulated to open and close a drawer.  The action space contains the end-effector positions in 3D.  We adopt MetaWorld reward functions and success metrics.  See supplementary materials for details.

\noindent
{\bf Testing.} We test the agent on surfaces of different textures which, unlike the grid texture used for training,  come from photos instead of from simulations.  
These tasks are extremely challenging for two reasons: 1) The agent has never seen any realistic textures during training; 2) Each texture also has a different color, so the agent needs to handle both color change and texture change at the same time.

\noindent
{\bf Texture background results.}
\label{DrawerWorldResults}
Table \ref{table:drawer} shows that our VAI outperforms PAD \cite{hansen2020self} and SAC \cite{haarnoja2018soft} significantly in all the test environments.  In particular, for 5 out of 6 textures such as blanket, metal, and wood, SAC and PAD have 0\% success rate, whereas VAI performs far better at 85\%, 98\%, and 70\% respectively.  In the training grid environment, PAD
 performs worse than SAC, consistent with \cite{hansen2020self} on the DeepMind Control benchmark, whereas our VAI is on-par or slightly better than SAC, suggesting that we are not gaining texture adaptability at the cost of losing training performance.

CNNs' sensitivity to textures poses a big challenge for visual adaptation. {\bf 1)} SAC adapts to unknown test environments with augmentations at the training time.  Since textures are not used during training, SAC breaks down during texture testing.  {\bf 2)} PAD has to change its feature encoder a lot in order to adapt to never-seen textures at the time time, shifting the feature distribution.  However, PAD assumes an invariant feature distribution and, therefore, does not fine-tune the control part of the policy network at the test time, which causes the vision-RL pipeline to break down.

\subsection{Ablation Studies}
\label{ablation}

\tabAblation{!t}

\noindent
We evaluate four ablated variants of our methods on the DrawerOpen task:
\begin{enumerate}[leftmargin=*,itemsep=-2pt,topsep=0pt]
\item SAC, a base  universal policy learning model
\item Method 1 $+$ RL with image augmentations, equivalent to Domain Randomization;
\item Method 2 $+$ Visual invariance module trained without augmentations: $\mathcal{T}_f, \mathcal{T}_b$ are identity functions;
\item Method 3 $+$ We apply the augmentations in Section \ref{section:visual_invariance} on $\mathcal{T}_f, \mathcal{T}_b$, for greater adaptability.
\end{enumerate}

Table \ref{table:ablation} shows that while all the methods perform well in the training environment, they adapt poorly to realistic textures except the last one. These results suggest that adding visual augmentations during RL or to the entire image as a whole is insufficient; providing a clean observation for RL agents with foreground clues adds significant robustness to vision-based RL.

\section{Summary}

We propose a fully unsupervised method to make vision-based RL more generalizable to unknown test environments.  While existing methods focus on learning a universal policy, we focus on learning universal foreground vision.

We learn to extract foregrounds with unsupervised keypoint detection, followed by unsupervised visual attention to remove model bias and generate a foreground mask. We then train a model to reconstruct the clean foreground mask from noise-augmented observations. 

We propose an additional challenging DrawerWorld benchmark, which trains manipulation tasks on grid and tests on texture environments.  Existing methods fail due to CNN's sensitivity to textures, yet our model with foreground extraction and strong generic augmentation is robust to never-seen textures without sacrificing training performance.

Our method significantly advances the state-of-the-art in vision-based RL, demonstrating that it is not only possible to learn domain-invariant vision without supervision, but freeing RL from visual distractions also improves the policy.

\vspace{5pt}
\noindent{\bf Acknowledgments.} This work was supported, in part, by Berkeley Deep Drive.

\def\tabEnvDesc#1{
\begin{table}[#1]
    \tablestyle{1.2pt}{1.1}
    \centering
    \begin{tabular}{l||p{0.8\linewidth\relax}}
    \shline
    Environment & Descriptions \\
    \shline
    Walker & A planar walker which encourages an upright torso and minimal torso height in the ``stand" task. In ``walk" task forward velocity is also encouraged. \\
    \hline
    Cartpole & A pole tied to a cart at its base, with forces applied to the base. ``swingup" task requires the pole to swing up from pointing down while ``balance" task requires the pole to balance to be upright. \\
    \hline
    Ball in cup & A ball attached to a cup, with forces applied to the cup to swing the ball up into the cup in the ``catch" task. \\
    \hline
    Finger & A finger is asked to rotate a rectangular body on a hinge. The top of the body needs to overlap with the object in ``turn\_easy" task and the body needs to rotate continuously in the ``spin" task.  \\
    \hline
    Cheetah & An animal with two feet which is asked to run in the ``run" task. \\
    \hline
    Reacher & A planar reacher with two links connected with a hinge in a plane with a random target location. In the ``easy" task, the reacher is asked to reach the object location. The ``hard" task is unused in our evaluation since it was not adapted by \cite{hansen2020self}. \\
    \shline
    \end{tabular}\vspace{-2pt}
    \caption{Descriptions for each environment in DeepMind Control suite.}
    \label{table:env_desc} 
\end{table}
}

\def\tabKeypointExpertiments#1{
\begin{table}[#1]
    \tablestyle{6pt}{1.1}
    \centering
    \begin{tabular}{p{0.55\linewidth\relax}||l}
    \shline
    RL Observations & Cumulative Reward \\
    \shline
    Joint Positions, Velocity, Torso Height from the Environment & 969\textpm2 \\
    \hline
    Joint Positions from the Environment & 935\textpm3 \\
    \hline
    \hline
    Keypoints Extracted with KeyNet & 709\textpm3 \\
    \hline
    VAI on Training Environment & 889\textpm3 \\
    \shline
    \end{tabular}\vspace{-2pt}
    \caption{Cumulative rewards on \textit{Walker, walk} task with 1) joint positions, velocity, and torso height from the environment as observations; 2) joint positions from the environment as observations; 3) keypoints extracted by KeyNet from images; 4) The proposed method VAI. The first two use the ground truth information, which is not accessible during real-world deployment, and serve as upper bounds. For experiment 2, 3, and 4, we use stack of 3 frames as input for the RL agent to infer the velocity since velocity information is missing. Since walker is a planar environment (the walker will not lean towards to away from the screen), the extracted keypoints should roughly correspond to positions from the significant parts of the walker body. The gap between experiments indicate that a limited number of keypoints from KeyNet on its own is not a sufficiently informative or accurate source for observations for an RL agent, which is in accord with our visualization in the main text about the keypoints' temporal inconsistency.}
    \label{table:keypoint_expertiments}
\end{table}
}

\def\figSamplesDMControl#1{
\begin{figure*}[#1]
\centering
\vspace{-12pt}
\includegraphics[width=1.0\linewidth]{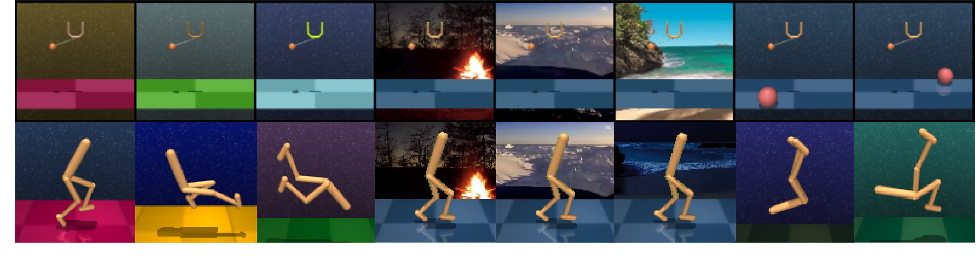}\vspace{-2pt}
\caption{Samples in evaluation environments in DeepMind Control. The samples in the first row are from \cite{hansen2020self}.}
\label{fig:sampled_dm_control}
\end{figure*}
}

\def\figSpeedMem#1{
\begin{figure}[#1]
\centering
\vspace{-12pt}
\includegraphics[width=1.0\linewidth]{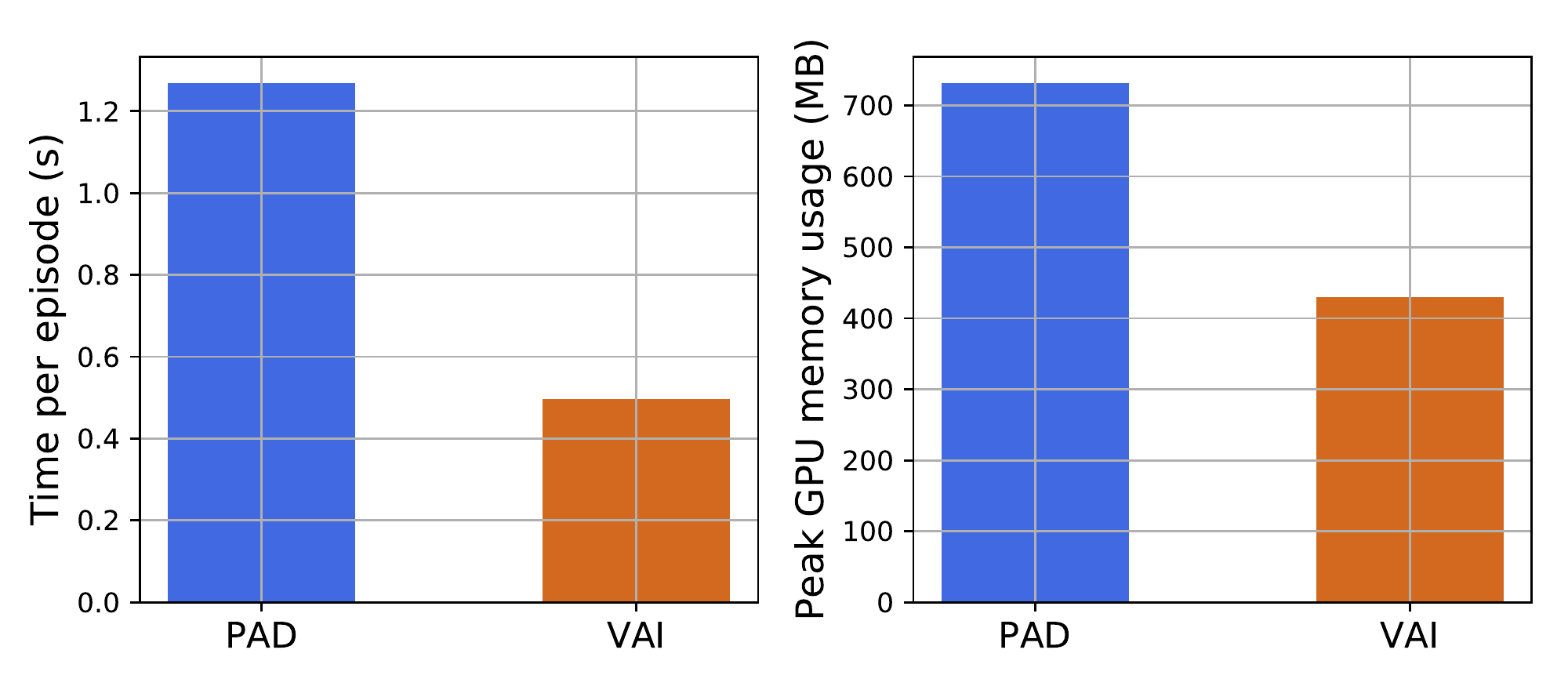}\vspace{-2pt}
\caption{Comparisons on the mean time per episode and GPU memory occupancy at evaluation time for \textit{DrawerClose} task in DrawerWorld between current state-of-the-art method PAD \cite{hansen2020self} and the proposed method VAI. VAI is more than 2 times faster than PAD during testing time and requires $\sim$40\% less GPU memory usage.
Both methods are evaluated with exactly the same backbone network. 
We take the mean of 10 runs for the latency comparison.
Memory usage is obtained with \texttt{torch.cuda.max\_memory\_allocated}.
}
\label{fig:speed_memory}
\end{figure}
}

\def\figMaskComparisons#1{
\begin{figure*}[#1]
\centering
\vspace{-12pt}
\includegraphics[width=1.0\linewidth]{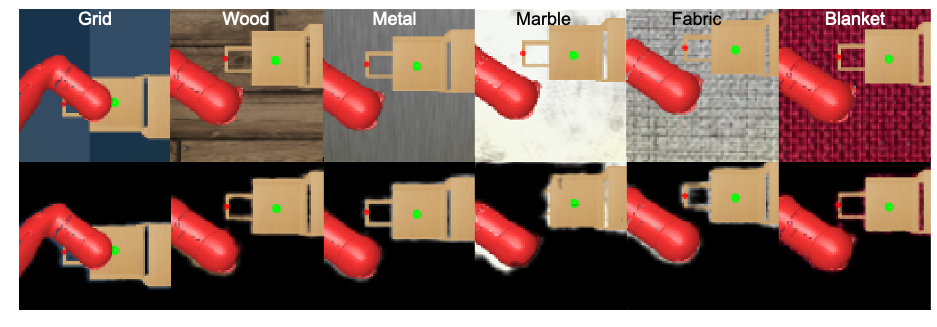}\vspace{-2pt}
\caption{Samples from DrawerWorld, \textit{DrawerClose} task with their corresponding observation processed by the adapter module. The Grid task is the training task for the adapter. All the observations use the same adapter for a fair comparison.}
\label{fig:mask_comparisons}
\end{figure*}
}

\def\tabDistractor#1{
\begin{table*}[#1]
    \tablestyle{5pt}{1.1}
    \centering
    \begin{tabular}{l||llllll}
    \shline
    Distracting objects & SAC \cite{haarnoja2018soft}  & DR \cite{hansen2020self}    & PAD \cite{hansen2020self}    & VAI & \multicolumn{2}{c}{$\Delta$ with SOTA} \\
    \shline
    Cartpole, swingup  & 815 \textpm 60  & 809 \textpm 24  & 771 \textpm 64  & \textbf{891\textpm0} & {\color{green(pigment)} \bf +76} &($\uparrow$9\%) \\
    Cartpole, balance  & 969 \textpm 20  & 938 \textpm 35  & 960 \textpm 29  & \textbf{993\textpm0} & {\color{green(pigment)} \bf +24} &($\uparrow$2\%) \\
    Ball in cup, catch & 177 \textpm 111 & 331 \textpm 189 & 545 \textpm 173 & \textbf{956\textpm4} & {\color{green(pigment)} \bf +420} &($\uparrow$77\%) \\
    Finger, spin       & 652 \textpm 184 & 564 \textpm 288 & \textbf{867 \textpm 72}  & 805\textpm3  & {\color{red} \bf -62}   &($\downarrow$7\%) \\
    Finger, turn\_easy & 302 \textpm 68  & 165 \textpm 12  & 347 \textpm 48  & \textbf{389\textpm18} & {\color{green(pigment)} \bf +42} &($\uparrow$12\%) \\
    \hline
    \it average       &\it 583  &\it 561  &\it 698 & \textbf{\textit{806}} &\it {\color{green(pigment)} \textit{\textbf{+108}}} &\textit{($\uparrow$15\%)} \\
    \shline
    \end{tabular}\vspace{-4pt}
    \caption{Comparisons with state-of-the-arts on testing environments with distracting objects. Mean and std of cumulative rewards across 10 random seeds and 100 random episode initializations per seed are compared for each task. The settings are similar to the ones in the randomized color experiments.}
    \label{table:distracting} 
\end{table*}
}

\section{Additional Environment Descriptions}
\subsection{DeepMind Control}

We wrote a short description for each environment in DeepMind Control suite \cite{tassa2020dmcontrol} in Table \ref{table:env_desc} to further introduce the environment.

\tabEnvDesc{!h}

We also provide samples for the evaluation environments designed by \cite{hansen2020self} in Fig. \ref{fig:sampled_dm_control}.

\figSamplesDMControl{!t}

\subsection{DrawerWorld}
We propose the DrawerWorld, a benchmark with observations in pixels, based on MetaWorld \cite{yu2019meta} to enable the agent to work in an environment close to real-life scenarios. There are two tasks in DrawerWorld, which are DrawerOpen and DrawerClose. These tasks ask a Sawyer robot to open and close a drawer, respectively. 

The multi-component reward function $R$ is a combination of a reaching reward $R_{\text{reach}}$ and a push reward $R_{\text{push}}$ as follows:
\begin{equation}
\begin{split}
        \hspace{-18pt} R &= R_{\text{reach}} + R_{\text{push}} \\
        &= -||h-p||_2 + \mathbb{I}_{||h-p||_2<\epsilon}\cdot c_1\cdot \text{exp}\{||p-g||^2_2/c_2\}
\end{split}
\end{equation}
where $\epsilon$ is a small distance threshold and is set as 0.08 by default, $p \in \mathbb{R}^3$ be the object position, $h \in \mathbb{R}^3$ be the position of the robot’s gripper, and $g \in \mathbb{R}^3$ be goal position. $c_1=1000$ and $c_2=0.01$ for all tasks in DrawerWorld benchmark.

The goal of distraction-robust RL is to learn a task-conditioned policy $\pi(a|s,z)$, where $z$ indicates an encoding of the task ID, and in this case, different task IDs have different drawer positions. This policy should maximize the average expected return from the task distribution $p(\mathcal{T})$, given by $\mathbb{E}_{\mathcal{T}\sim p(\mathcal{T})}[\mathbb{E}_{\pi}[\sum_{t=0}^T\gamma^tR_t(s_t, a_t)]]$. The success metric, which is evaluate the agent in evaluation time, is described by $\mathbb{I}_{||p-g||_2<\epsilon}$, where $\epsilon$ is set to 8cm.
The difference between training and evaluation time is the texture and color of the table cloth. Image samples of textures that we use are provided in the main text.

\section{De-noise with Past Averages}

Since our adapter model works on each frame separately without any assumption on temporal continuity of consecutive frames, our adapter works exactly the same on videos as on fixed backgrounds and is not affected by drastic changes in the background such as flashes of light. However, in some environments where the assumption of temporal continuity holds, i.e. with a relatively slow-moving background, we may make use of this assumption to better de-noise the observations before passing the them into the adapter.

We exploit the assumption here by keeping a mean of past observations ${\bf{o}}_{\text{mean}}=\frac{1}{t}\sum^{t}_{i=1} {\bf{o}}_{i}$ and subtract the mean from observation ${\bf{o}}_{t}$ and compute the observations after de-noise with the formulation: 
This de-noise step happens before the observation is sent into CNN (adaptor), formulated as:
\begin{equation}
    {\bf{o}}_{\text{de-noised}} = \text{filter}({\bf{o}}_{t} - \alpha {\bf{o}}_{\text{mean}}, \epsilon) + \alpha {\bf{o}}_{\text{mean\_color}}
\end{equation}
where ${\bf{o}}_{\text{mean\_color}}$ is the mean of ${\bf{o}}_{\text{mean}}$ in spatial dimensions, $\text{filter}$ is a function that sets the part with value less than $\epsilon$ to 0 to remove some noise, and $\alpha \in [0, 1]$ is the strength in noise removal.

\textit{This is completely optional}, and since it makes use of an additional assumption, they are only used in the cartpole and ball in cup in experiments with background videos and experiments with distracting objects. In addition, we observe that with Places dataset as augmentation, the model is robust enough without this trick, so we disable it for all models in the training of which Places dataset is used.

\section{Memory Usage and Speed Comparisons}
\label{section:speed_memory}
\figSpeedMem{!h}

\figMaskComparisons{!th}

From Fig. \ref{fig:speed_memory}, it seems that VAI is about 3 times as fast as PAD in terms of the evaluation time in each episode and requires substantially less GPU memory than PAD. This is largely due to the fact that PAD trains the encoder network at evaluation time with back-propagation, which not only requires the intermediate results to be saved in GPU memory but also requires backward computation to update the model parameters, which consumes both time and memory space. Although VAI has an extra adapter module, the computation and memory it takes are much less than the ones required by backward computation and storing intermediate results. According to the requirements of computational resources in terms of speed and memory, our method is more suitable for robots powered by battery and edge inference devices than PAD from this point of view.

\tabKeypointExpertiments{!th}

\section{Further Experiments with Raw Keypoints}

To investigate the question of whether raw keypoints extracted from image observations by KeyNet are able to contribute to effective learning of useful behaviors, we set up experiments based on the \textit{Walker, walk} task in DeepMind Control and list the outcomes in Table \ref{table:keypoint_expertiments}.

We first train an agent with the state-based observation provided by the environment, a 24-dimensional tensor, which includes positions of the joint, velocity, and walker's torso height. We do not stack frames for this experiment. This experiment indicates an upper bound that our agent is able to achieve in this environment. However, to make a fair comparison with other experiments, where velocity and torso height information is not directly provided, we also remove these parts from our observation, leaving only the positions of the joint as observation in the second experiment. Thus, in the second experiment, the agent directly reads a 14-dimensional tensor per frame from the environment based on the position of each joint. The third experiment is conducted with the RL agent reading a tensor that contains the $(x,y)$ coordinates of 24 keypoints. The keypoints are from a KeyNet which reads an image input. The KeyNet is pre-trained with transporter. In the last experiment, we run VAI, with the adapter module trained from the same KeyNet that is used in the experiment above, on the training environment, although it is able to adapt to other environments as well and thus is more general. To infer velocity information, we stack the observations for three frames for experiments 2, 3, and 4. We run both experiments for 500k steps. We compare their efficiency by evaluating the agent in training environment 10 times with 10 seeds.

According to the performance of these RL agents in the training environment, keypoints on its own do not capture all the information needed by the RL agent accurately. This will be even worse if the agent is evaluated in a different environment is has never seen before, since KeyNet itself does not come with the ability to adapt, although the keypoints it generates are not supposed to carry domain-specific or distraction information. Using keypoints information along with image features as well as history observations may help, as illustrated in \cite{kulkarni2019unsupervised} and described in the main text, but it will add greatly to the complexity of the RL framework. What's more, agents may need information other than what keypoints provide. For example, keypoints do not carry the shape, size, and color information, which may be of paramount importance in certain tasks. Furthermore, since KeyNet allocates an output dimension for each keypoint, the number of parameters as well as computation time scales linearly with the number of keypoints, which prohibits adding a large number of keypoints to compensate the effect of temporal inconsistency or to capture complicated observations. In contrast, since KeyNet is not used in getting adapted observations in our method, the speed and number of parameters of our RL agent, including the adapter, at evaluation time are not affected by the number of keypoints used to generate ground-truth, which allows our method to scale to complicated environments with many moving parts without losing efficiency.

\section{Visualizations of the Observation Adapter}
How to make sure that RL will adapt to a certain setting that is different from training setting is still an open problem. Our method opts to work on observation-space. In contrast, PAD works on an intermediate encoder feature space. Our method is much easier to visualize and debug since humans are able to directly understand the quality of adapted observations while it is really difficult to understand what happens in the feature space.

To give examples on how to assess whether an adapter works on a certain environment easily and to illustrate our performance in a visual way in evaluation environments, we gathered 6 pairs of raw samples and samples processed by the adapter in \textit{DrawerClose} task from the same adapter in Fig. \ref{fig:mask_comparisons}. As can be seen from the examples, the adapter model differentiates most of the evaluation environments well, with the exception of the marble environment, which the adapter confuses parts of the foreground and background such as the handle and the patches around the actuator, probably due to the fact that the reflected light on the actuator has a similar white color to the color of background. This indicates why our model performances worse in marble environment, as illustrated in the experiment section in the main text, and, in real-life applications, means that the adapter needs to be re-trained or fine-tuned with observations from similar environments, or if this is not applicable, with augmentation specially-designed to handle this case. We leave the question of handling adapter fine-tuning and re-training to later research.

This visualization has a large impact on the real-world applications of our method: with only a few observations from an intended deployment environment, one could easily visualize and assess whether our method will adapt to such environment. This does not require any ability to run the policy in the dynamics, nor does it require reward functions or consecutive observations which may be difficult to obtain from deployment environments in real-world applications. We strongly believe that this simple assessment provides a direction for future research in explainable, adaptable, and generalizable reinforcement learning and will present great benefit to potential applications of reinforcement learning.

{\small
\bibliographystyle{ieee_fullname}
\bibliography{egbib}
}

\end{document}